\documentclass{article}
\usepackage{spconf,amsmath,graphicx}

\usepackage{amsmath,amsfonts}
\usepackage{algorithmic}
\usepackage{algorithm}
\usepackage{array}
\usepackage[caption=false,font=normalsize,labelfont=sf,textfont=sf]{subfig}
\usepackage{textcomp}
\usepackage{stfloats}
\usepackage{url}
\usepackage{verbatim}
\usepackage{graphicx}
\usepackage{cite}
\usepackage{amsmath,amsfonts}
\usepackage{mathrsfs}
\usepackage{stfloats}
\usepackage{diagbox}
\usepackage{bm}
\usepackage{booktabs,multirow}
\usepackage{arydshln} 
\usepackage{listings}%
\usepackage{enumitem}

\title{SSTA: Salient Spatially Transformed Attack}
\name{Author(s) Name(s)\thanks{Thanks to XYZ agency for funding.}}
\address{Author Affiliation(s)}
\name{Renyang Liu$^{1,2,\dagger}$ \qquad Wei Zhou$^{1}$ \qquad Sixin Wu$^{1}$ \qquad Jun Zhao$^{2}$ \qquad Kwok-Yan Lam$^{2}$}
  
\address{$^{1}$ Yunnan University \\
      $^{2}$ Nanyang Technological University\\
      $^{\dagger}$ This paper was accepted by ICASSP 2024.}

\begin{document}
%
\maketitle
\begin{abstract}
Extensive studies have demonstrated that deep neural networks (DNNs) are vulnerable to adversarial attacks, which brings a huge security risk to the further application of DNNs, especially for the AI models developed in the real world. Despite the significant progress that has been made recently, existing attack methods still suffer from the unsatisfactory performance of escaping from being detected by naked human eyes due to the formulation of adversarial example (AE) heavily relying on a noise-adding manner. Such mentioned challenges will significantly increase the risk of exposure and result in an attack to be failed. Therefore, in this paper, we propose the Salient Spatially Transformed Attack (SSTA), a novel framework to craft imperceptible AEs, which enhance the stealthiness of AEs by estimating a smooth spatial transform metric on a most critical area to generate AEs instead of adding external noise to the whole image. Compared to state-of-the-art baselines, extensive experiments indicated that SSTA could effectively improve the imperceptibility of the AEs while maintaining a 100\% attack success rate. 
\end{abstract}

%

\section{Introduction}
\label{sec:intro}
Deep neural networks (DNNs) are susceptible to AEs, which are crafted by subtly perturbing a clean input \cite{ijautcomp/XuMLDLTJ20}, especially for computer vision (CV) tasks, like image recognition. The critical point to carry out adversarial attacks on CV models is how to generate AEs with attack success rate and high imperceptibility. Various methods have been proposed to build AEs; among them, most such attacks are crafting AEs in optimizing noise and adding noise manner.

Although most existing attacks can obtain a high success rate by adding noise to the original image, they are not ideal in terms of imperceptibility and similarity since the added perturbations are not harmonious with the clean image \cite{sp/Carlini017,iclr/MadryMSTV18}. To address these issues, researchers have proposed various works. Some methods try to generate AEs in a non-noise addition way, such as the spatial transform-based attack, which crafts AEs by changing the specific pixel's position \cite{iclr/XiaoZ0HLS18, mm/AydinSKHT21}. Even though these methods ensure the adversarial perturbations are more harmonious with the clean counterparts, the imperceptibility is still weak because they disturb the entire image. In most cases, people can easily distinguish the AEs generated by these methods through the naked eyes. 

To improve the concealment of AEs, we formulate the issue of synthesizing AEs beyond additive perturbations and propose a novel non-addition attack method called \textbf{SSTA}. More specifically, SSTA uses spatial transformation techniques \cite{nips/JaderbergSZK15} based on the salient region of the image to generate AEs, rather than directly adding well-designed noise to the benign image. The spatial transform technique can learn a smooth flow field for each pixel's new locations to optimize an eligible AE. To further ensure the concealment and image quality, we constraint the optimized flow field $\bm{f}$ (the transform metric) by limiting it with a small dynamic flow budget $\xi$. 

Extensive experiments on ImageNet datasets indicate that the proposed SSTA can make AEs more inconspicuous while maintaining high attack performance. Besides, evaluation results on many metrics involve similarity and image quality showing that our AEs are more similar to their benign counterparts and preserved the vivid details. The main contributions could be summarized as follows:
\begin{itemize}[leftmargin=*, itemsep=1pt, topsep=1pt, parsep=0pt]
    \item We formulate the imperceptible AE by spatial transform operations in the local salient region, which are extracted by object detection method rather than in a noise-adding manner. 

    \item To balance the attack performance and the concealment of the generated AEs, we propose a dynamic strategy to update the extracted critical region and flow budget $\xi$ associated with the number of optimizations increases.  
    
    \item Comparing with the state-of-the-art imperceptible attacks, experimental results on various victim models show our method's superiority in synthesizing AEs with the attack ability, invisibility, and image quality and guarantee the AEs' similarity to the original image.
\end{itemize}

The rest of this paper is organized as follows. In Sec. \ref{sec:methodology}, we provide the details of the proposed SSTA framework. The experiments are presented in Sec. \ref{Sec:experiments}, with the conclusion drawn in Sec. \ref{Sec:conclusion}.

\begin{figure*}[htp]
      \centering
      \includegraphics[width=0.80\textwidth]{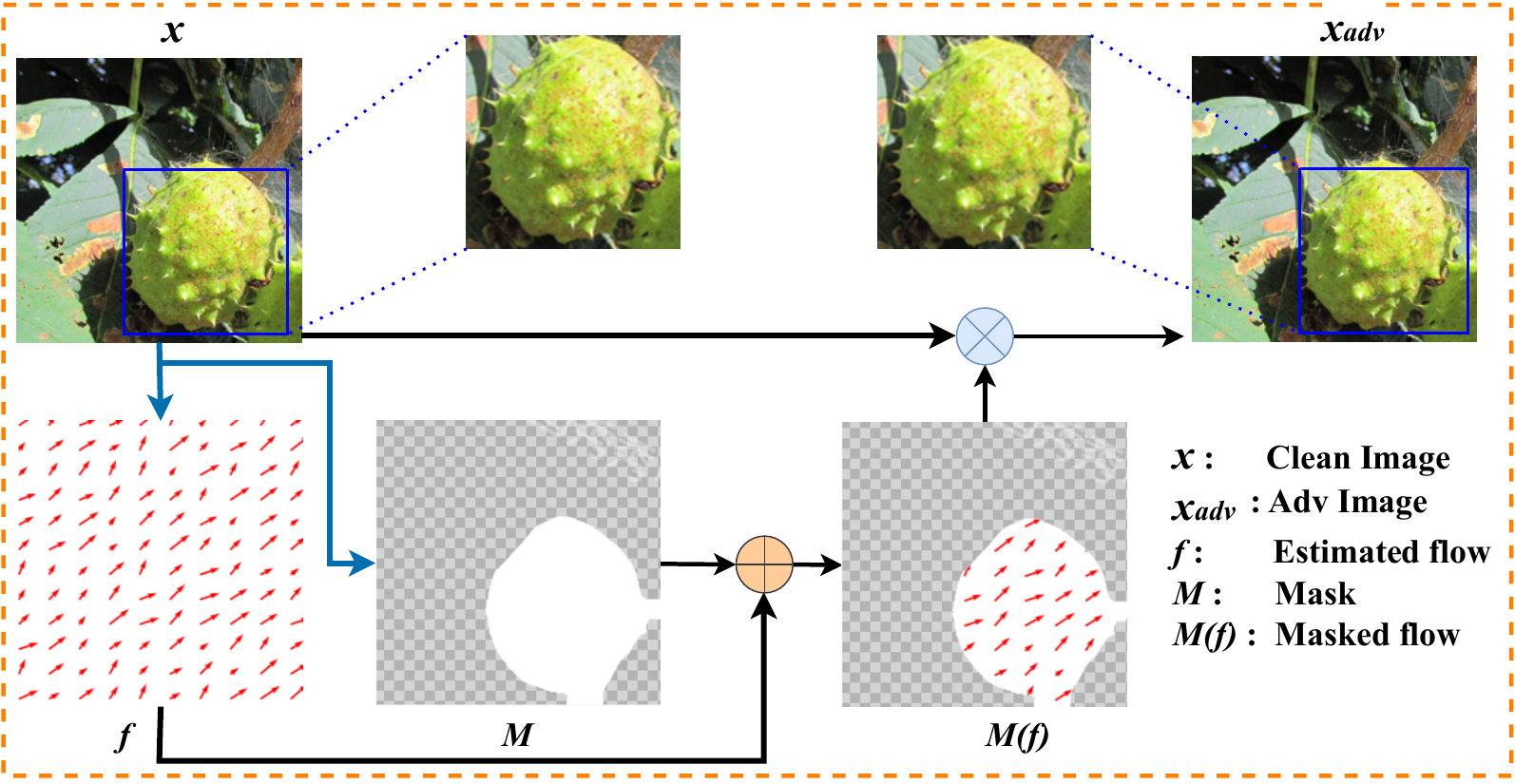}
      \caption{Overview of SSTA, where $\oplus $ represents applying Mask $M$, and $\otimes$ represents the spatial transformation operation.}
      \label{fig:framwork}
      \vspace{-0.4cm}
\end{figure*}
\section{Methodology}
\label{sec:methodology}

\subsection{Problem Definition}
Given a well-trained DNN classifier $ \bm{\mathcal{C}} $ and an input $ \bm{x} $ with its corresponding label $ y $, we have $ \bm{\mathcal{C}}(\bm{x})=y $. The AE $ \bm{x}_{adv} $ is a neighbor of $ \bm{x} $ and satisfies that $ \bm{\mathcal{C}}(\bm{x}_{adv}) \neq y $ and $ \left \| \bm{x}_{adv}-\bm{x} \right \|_p \leq \epsilon  $, where the $ \bm{L}_p $-norm is used as the metric function and $ \epsilon $ is usually a small noise budget. With this definition, the problem of finding an AE becomes a constrained optimization problem:
\begin{equation}
      \label{eq:eq1}
      \bm{x}_{adv}= 
\underset{\left \| \bm{x}_{adv}-\bm{x} \right \|_p \leq \epsilon}{\mathop{arg max}\mathcal{L}} ( \bm{\mathcal{C}}(\bm{x}_{adv}) \neq y),
\end{equation}
where $ \mathcal{L} $ stands for a loss function that measures the confidence of the model outputs.

Previous works craft an AE $\bm{x}_{adv}$ by adding $ \bm{L}_p $-norm constrained noise $\delta$ to the clean image $\bm{x}$ as
\begin{equation}
    \label{eq:eq2}
    \bm{x}_{adv} = \bm{x}+\delta, \  s.t. \  \left \| \delta \right \|_p \leq \epsilon.
\end{equation}
Different from this, in this paper, we combine the salient object extraction and the spatial transform techniques to build the imperceptible AE $\bm{x}_{adv}$. As illustrated in Fig.~\ref{fig:framwork}, the proposed salient spatially transformed attack framework can be divided into two stages: the first stage is to obtain a salient region, we call it mask $\bm{M}(\cdot) $; the other one is to calculate the flow field $ \bm{f} $. Subsequently, we can formulate the AE $\bm{x}_{adv}$ by applying the calculated flow field $ \bm{f} $ to the clean image's salient area $ \bm{M}(\cdot) $.

\subsection{Salient Region Extraction}
\label{subsec:mask_region}
\begin{figure}[htp]
      \centering
      \includegraphics[width=0.47\textwidth]{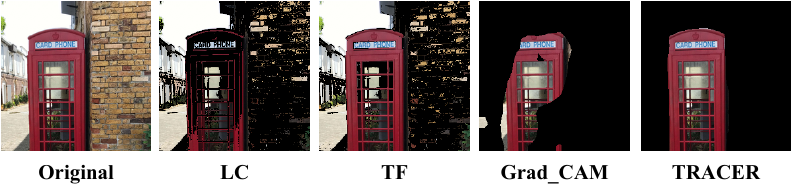}
      \caption{The extracted area by different methods.}
      \label{fig:mask} 
\end{figure}

In this paper, we use the salient detection method TRACER \cite{aaai/LeeSH22}, which can efficiently detect salient objects in images, to extract the critical area mask $\bm{M}(\cdot)$. In preliminary experiments, we also tried other area extraction methods like LC \cite{mm/ZhaiS06}, FT \cite{cvpr/AchantaHES09}, and Grad-CAM \cite{iccv/SelvarajuCDVPB17,prl/DengZ19}, but found TRACER \cite{pr/QinZHDZJ20} is more suitable because it can efficiently detect salient objects in an image and return their corresponding regions, the results are showing in Fig.~\ref{fig:mask}. 

Moreover, TRACER can return several regions $ r_\tau(\tau=0, ..., 255)$ with various scales depending on different thresholds when extracting salient areas. These regions will be helpful to the downstream tasks, such as image segmentation and background removal. In our work, we first take the region with a high threshold $\tau$, such as $\tau=250$, as the region mask $\bm{M}(\cdot)$. Then, in generating AEs, if the current attack is unsuccessful after pre-set iterations, the $\bm{M}(\cdot)$ will be updated by decreasing the threshold $\tau$.

\subsection{Adversarial Example Generation}
\label{subsec:adversarial}
After computing the mask region $\bm{M}(\cdot)$, we subsequently utilize the spatial transform method to build AEs in a non-noise additional way. The spatial transform techniques using a flow field matrix $\bm{f}=[2, h, w]$ to transform the original image $x$ to $ x_{st}$ \cite{iclr/XiaoZ0HLS18}. Specifically, assume the input is $x$ and its transformed counterpart $ \bm{x}_{st}$, for the $i$-th pixel in $\bm{x}_{st}$ at the pixel location $ (u_{st}^i,v_{st}^i) $, we need to calculate the flow field matrix $\bm{f}_i = (\Delta u^i, \Delta v^i) $. So, the $i$-th pixel $ \bm{x}^i$'s location in the transformed image can be indicated as:
\begin{equation}
      \label{eq:eq3}
      (u^i,v^i) =(u_{st}^i + \Delta u^i ,v_{st}^i+ \Delta v^i ).
\end{equation}

To ensure the flow field $ \bm{f} $ is differentiable, the bi-linear interpolation \cite{nips/JaderbergSZK15} is used to obtain the 4 neighboring pixels' value surrounding the location $(u_{st}^i + \Delta u^i, v_{st}^i+ \Delta v^i )$ for the transformed image $\bm{x}_{st}$ as:
\begin{equation}
      \label{eq:eq4}
      \bm{x}_{st}^i=\sum_{q \in \bm{N} (u^i,v^i)}\bm{x}^q(1-|u^i-u^q|) (1-|v^i-v^q|),
\end{equation}
where $ \bm{N} (u^i,v^i) $ is the neighborhood, that is, the four positions (top-left, top-right, bottom-left, bottom-right) tightly surrounding the target pixel $ (u^i,v^i) $. In adversarial attack settings, the calculated $\bm{x}_{st}$ is the final AE $\bm{x}_{adv}$. Once the $ f $ has been computed, we can obtain the $\bm{x}_{adv}$ by combining $\bm{M} (\cdot)$ and flow field $\bm{f}$, which is given by:
\begin{multline}
    \label{eq:eq5}
    \bm{x}_{adv}= clip(\bm{M}(\sum_{q \in \bm{N} (u^i,v^i)}\bm{x}^q(1-|u^i-u^q|) (1-|v^i-v^q|)) \\ 
    + (\bm{x}-\bm{M(\bm{x}})),0,1),
\end{multline}
where $\bm{M}(\bm{x})$ represents the salient region while the $\bm{x}-\bm{M}(\bm{x})$ indicates the area out of the salient region.

In practice, we regard the problem of calculating flow field $\bm{f}$ as an optimization task. In this paper, we use the AdamW to optimize flow $\bm{f}$.

\subsection{Objective Functions}
Taking the attack success rate and visual invisibility of the generated AEs into account, we divide the objective function into two parts, where one is the adversarial loss and the other is a constraint for the flow field. Unlike other flow field-based attack methods, which constrain the size of the flow field by the flow loss proposed in \cite{iclr/XiaoZ0HLS18}, in our method, we use a dynamically updated flow field budget $ \xi $ (a small number, like $1*10^{-2}$) to regularize the flow field $f$. For adversarial attacks, the goal is making $\bm{\mathcal{C}}(\bm{x}_{adv}) \neq y$. We give the objective function as:
\begin{multline}
    \label{eq:eq6}
     \mathcal{L}_{adv} (\bm{x},y,\bm{f})= max[\bm{\mathcal{C}}(\bm{x}_{adv})_{y}-\underset{k\neq y}{max}\bm{\mathcal{C}}(\bm{x}_{adv})_{k},k],\\
     s.t. \| \bm{f} \| \leq \xi.
\end{multline}

\section{Experiments}
\label{Sec:experiments}
\subsection{Settings}

\textbf{Dataset:} We verify the performance of our method on the development set of ImageNet-Compatible Dataset, a subset of ImageNet-K, which consists of 1,000 images with a size of 299×299×3. And we resized the image to 224x224x3 to adopt the victim models.

\textbf{Models:} We use the PyTorch pre-trained model as the victim models, including VGG-19 \cite{corr/SimonyanZ14a}, ResNet-50 \cite{cvpr/HeZRS16}, DenseNet-121 \cite{cvpr/HuangLMW17}, ViT-16 \cite{iccv/LiuL00W0LG21} and Swin\_B \cite{iclr/DosovitskiyB0WZ21}.

\textbf{Baselines:} The baselines include the stAdv \cite{iclr/XiaoZ0HLS18}, Chroma-Shift \cite{mm/AydinSKHT21} and AdvDrop \cite{iccv/DuanCNYQH21}. 

\textbf{Metrics:} 
Unlike the pixel-based attack methods, which always use $ L_p $-norm to evaluate the AEs' perceptual similarity to its corresponding benign image. The AEs generated by spatial transformation always use other metrics referring to image quality. To be exact, we use the following perceptual metrics to evaluate the AEs generated by our method, including LPIPS \cite{cvpr/ZhangIESW18}, DISTS \cite{pami/DingMWS22}, FID, MSE, UQI \cite{spl/WangB02}, SCC \cite{scc}, PSNR \cite{tip/SheikhSB06}, VIPF \cite{icassp/SheikhB04}, SSIM, and NIQE to evaluate the difference between the generated AEs and their benign counterparts and the image quality of these AEs. 

\subsection{Attacking Performance}
We investigate the performance of the proposed method in attacking various image classifiers. The results are shown in Table. \ref{tab:attack}, we derive that SSTA can obtain the SOTA attack performance by only disturbing the minimal local area, i.e., the salient region, while other attacks need to distort the whole image. This demonstrates the superiority of our method.
\vspace{-12pt}

\begin{table}[ht]
\centering
\small
\renewcommand{\arraystretch}{1}
\setlength\tabcolsep{1.5pt}
\caption{Attack performance of baselines and SSTA.}
\label{tab:attack}
\begin{tabular}{cccccc}
\toprule
Methods     & VGG-19       & ResNet-50    & DenseNet-121 & VIT-16       & Swin\_B       \\
\midrule
stAdv        & \textbf{100} & \textbf{100} & \textbf{100} & \textbf{100} & \textbf{100}  \\
Chroma-Shift & 93.69        & 94.67        & 95.1         & 95.09        & 96.66         \\
AdvDrop      & \textbf{100} & 99.07        & \textbf{100} & 95.97        & 99.79         \\
SSTA         & \textbf{100} & 99.86        & \textbf{100} & \textbf{100} & \textbf{100}  \\
\bottomrule
\end{tabular}
\end{table}

\vspace{-12pt}

\subsection{Image Quality and Similarity}
We use diverse metrics involving image quality and similarity to assess the AEs' image quality and list the results in Table. \ref{tab:perceptual}, which indicated that the proposed method has the lowest LPIPS, DISTS, FID, and MSE (the lower is better) are 0.0038, 0.0091, 16.3876 and 2.1210, respectively, and has the highest UQI, SCC, PSNR, VIPF, SSIM, and NIQE (the higher is better), achieving 0.9998, 0.9890, 49.2397, 0.9487, 0.9987 and 43.9611, respectively, in comparison to the baselines. The results point out that the proposed method is superior to the existing imperceptible attacks.

\begin{figure}[htb]
      \centering
      \includegraphics[width=0.49\textwidth]{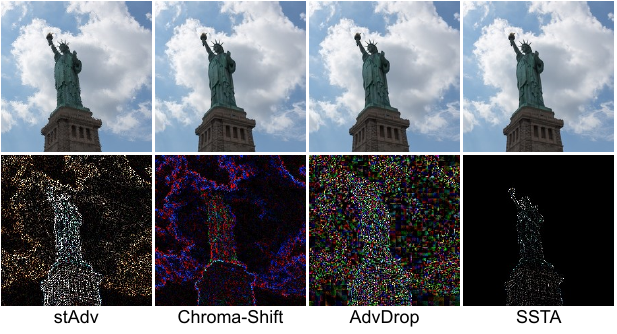}
      \vspace{-25pt}
      \caption{AEs and their corresponding perturbations.}
      
      \label{fig:visual_2}
      \vspace{-10pt}
\end{figure}

\begin{table}
\setlength{\abovecaptionskip}{-0pt} 
\setlength{\belowcaptionskip}{-0pt}
\caption{Perceptual distances were calculated on fooled examples by stAdv, Chroma-Shift and the proposed SSTA.}
\label{tab:perceptual}
\centering
\renewcommand{\arraystretch}{1}
\setlength\tabcolsep{3.5pt}
\begin{tabular}{ccccc} 
\toprule
Metrics & stAdv   & Chroma-Shift & AdvDrop & SSAT              \\ 
\midrule
LPIPS $\downarrow$  & 0.1595  & 0.0135       & 0.0956  & \textbf{0.0038}   \\
DISTS $\downarrow$  & 0.1524  & 0.0165       & 0.0678  & \textbf{0.0091}   \\
FID $\downarrow$    & 60.2464 & 88.8750      & 46.7813 & \textbf{16.3876}  \\
MSE $\downarrow$    & 95.7488 & 23.5399      & 17.0450  & \textbf{2.1210}   \\

\hdashline
UQI $\uparrow$    & 0.9925  & 0.9925       & 0.9952  & \textbf{0.9998}   \\
SCC $\uparrow$    & 0.6415  & 0.9623       & 0.6894  & \textbf{0.9890}   \\
PSNR $\uparrow$    & 29.8119 & 36.5651      & 36.1464 & \textbf{49.2397}  \\
VIFP $\uparrow$    & 0.5229  & 0.7644       & 0.6474  & \textbf{0.9487}   \\
SSIM $\uparrow$   & 0.9391  & 0.9771       & 0.9688  & \textbf{0.9987}   \\

NIQE $\uparrow$   & 33.3234 & 43.5860      & 39.8657 & \textbf{43.9611}  \\
\bottomrule
\end{tabular}
\vspace{-15pt}
\end{table}



To visualize the difference between the AEs generated by our method and the baselines, we also draw the adversarial perturbation generated by stAdv, Chroma-Shift, AdvDrop and the proposed method in Fig.~\ref{fig:visual_2}, the target model is pre-trained ResNet-50. The first row is the AEs and their corresponding noise of stAdv, Chroma-Shift, AdvDrop and our method, respectively. Noted that, for better observation, we magnified the noise by a factor of 30. From Fig.~\ref{fig:visual_2}, we can clearly observe that these baselines distort the whole image. In contrast, the AEs generated by our method are focused on the salient region and its noise is milder; they are similar to the original clean counterparts and are more imperceptible to human eyes. These results indicate that the AEs generated by the proposed method have better concealment and are not easily exposed.

\subsection{Further Human Perceptual Study}
\label{sec:humen_study}
This experiment is for subjective evaluation, i.e., in most cases, whether AEs generated based on SSTA are indistinguishable from their original samples. We argue that AEs generated by SSTA not only satisfy imperceptibility but are also inconspicuous to the human eye. To validate this claim, we compare AEs generated by SSTA with those generated by baseline methods. In our human perception study, we display the original image and the AEs on the computer screen and give each participant 100 seconds to judge every image. Empirically, 100 seconds is enough to decide and point out any visible distortion for the participants. We used the randomly sampled 50 images from the ImageNet dataset for this experiment. The participants are shown about 5 images, the left is always the clean image and its right side shows adversarial images generated by various methods (Maybe more than one) or the same clean image. Participants will be asked ``Are the images on the right the same as the left (the clean one)?'' and each participant will provide more than 50 annotations. For each image to be checked, we can zoom it as large as possible to provide convenience for participants to observe. 

\begin{figure}[!ht]
      \centering
      \includegraphics[width=0.48\textwidth]{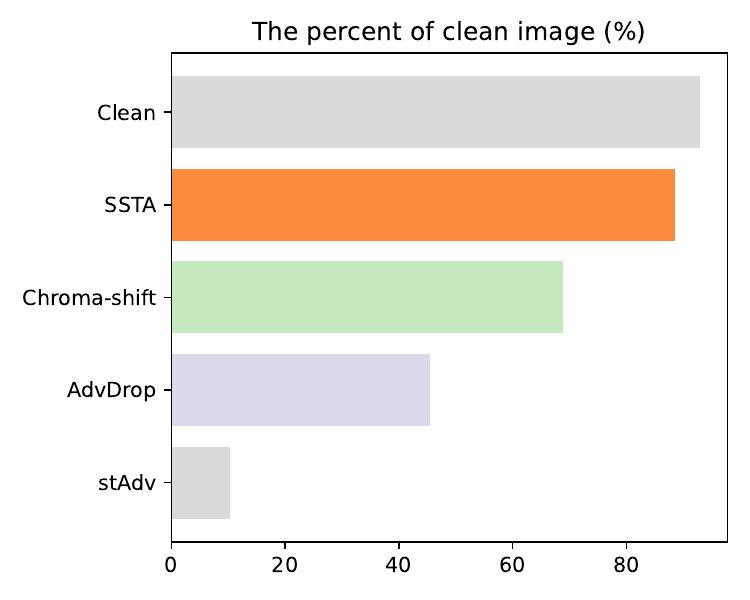}
      \vspace{-20pt}      
      \caption{Human perceptual study results where the bars signify the percentage of the participants answering ``The images are the same as the clean image".}
      \label{fig:human}
      \vspace{-15pt}
\end{figure}

A total of 20+ participants were involved in assessing whether the specific image was the same as the original clean image. For the sake of fairness, we put the clean images and adversarial images generated by different methods into the dataset to be checked together. These participants provided more than 1,000 annotations. As shown in Fig.~\ref{fig:human}, the AEs generated by SSTA are generally considered to be the same as the original images. 88.98\% of the annotations were considered unmodified, meaning that most participants could not distinguish the AEs generated by SSTA. Conversely, for AEs generated by baseline methods, participants were able to spot distortions more easily, more than 90\%, 55\% and 30\% of the total annotations have been picked out for stAdv, AdvDrop and Chroma-shift, respectively, indicating that the AEs generated by these baseline methods did not affect humans to identify objects in images correctly but very easy to find that they had tampered.

\section{Conclusions}
\label{Sec:conclusion}
In this paper, we present a novel non-noise additional method, called SSTA, which combines performing the spatial transformation in salient regions with the optimal flow field to synthesize AEs. Extensive experiments show that the proposed method is superior to the state-of-the-art methods in terms of prominent concealment and high image quality, and the generated AEs are indistinguishable by the human eyes. Benefitting from generating AEs without noise-adding, the proposed SSTA provides a new efficient way to evaluate the robustness of classifiers and enhance their performance using techniques like fine-tuning or adversarial training. Furthermore, the proposed approach can be used as a reliable tool to build more robust models.


\clearpage
\bibliographystyle{IEEEbib}
\bibliography{refs}

\end{document}